\documentclass[runningheads]{llncs}
\usepackage{graphicx}
\usepackage{bm}
\usepackage{amsmath} 
\usepackage{amssymb} 
\usepackage{mathtools} 
\usepackage{multirow}
\usepackage{booktabs}
\usepackage[colorlinks=true, linkcolor=red, citecolor=green, urlcolor=cyan]{hyperref}

\begin{document}

\title{GuidPaint: Class-Guided Image Inpainting with Diffusion Models}
\titlerunning{GuidPaint}
\author{Qimin Wang\inst{1}\and Xinda Liu\inst{1}\and Guohua Geng\inst{1}}
\authorrunning{Q. Wang et al.}
\institute{Northwest University, Xi'an, China\\
\email {qmwang@stumail.nwu.edu.cn},
\email{\{liuxinda, ghgeng\}@nwu.edu.cn}}
\maketitle              

\begin{abstract}
In recent years, diffusion models have been widely adopted for image inpainting tasks due to their powerful generative capabilities, achieving impressive results. Existing multimodal inpainting methods based on diffusion models often require architectural modifications and retraining, resulting in high computational cost. In contrast, context-aware diffusion inpainting methods leverage the model's inherent priors to adjust intermediate denoising steps, enabling high-quality inpainting without additional training and significantly reducing computation. However, these methods lack fine-grained control over the masked regions, often leading to semantically inconsistent or visually implausible content. To address this issue, we propose GuidPaint, a training-free, class-guided image inpainting framework. By incorporating classifier guidance into the denoising process, GuidPaint enables precise control over intermediate generations within the masked areas, ensuring both semantic consistency and visual realism. Furthermore, it integrates stochastic and deterministic sampling, allowing users to select preferred intermediate results and deterministically refine them. Experimental results demonstrate that GuidPaint achieves clear improvements over existing context-aware inpainting methods in both qualitative and quantitative evaluations. Our code is available at \url{https://github.com/wangqm518/GuidPaint}.

\keywords{Image Inpainting  \and Diffusion Models \and Classifier Guidance \and Training-Free \and Sampling Strategy \and Zero-Shot Learning.}
\end{abstract}
\begin{figure}
\includegraphics[width=\textwidth]{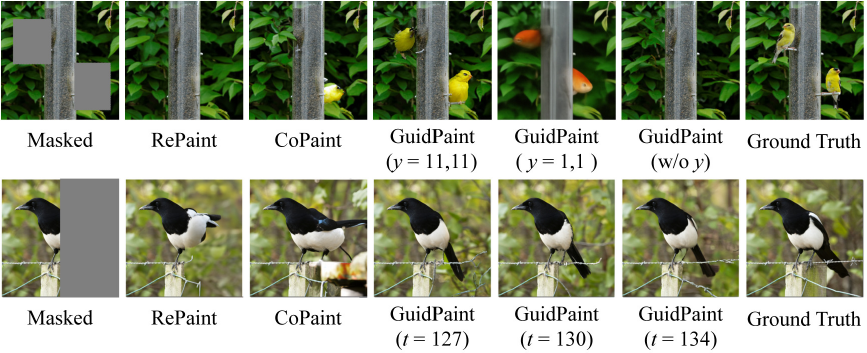}
\caption{GuidPaint (Ours) enables class-guided control over masked regions (top), while stochastic sampling generates visually plausible content (bottom), compared to context-aware inpainting methods. Here, class label $y$ denotes conditional diffusion model with classifier guidance (11: ``goldfinch", 1: ``goldfish"), and w/o $y$ indicates unconditional diffusion model without classifier guidance. Bottom: $y$=18 (``magpie"). } \label{fig1}
\end{figure}
\section{Introduction}
Image inpainting is a crucial task in computer vision, serving as both a subtask of image restoration and a foundation for image editing (e.g., object removal, object replacement, background replacement, style transfer). The goal of image inpainting is to generate semantically consistent and visually plausible content in unknown regions based on the context of known areas. In previous work, image inpainting methods have introduced and explored various deep generative models, \cite{lama,LSIC-GANs} are based on Generative Adversarial Networks (GANs) \cite{GAN},
\cite{rethinking} is based on Auto-Encoders (AEs) \cite{AutoEncoder}, \cite{VQ-VAE-inp,Semantic-inp-VAE} rely on Variational Auto-Encoders (VAEs) \cite{VAE}, and \cite{high-inp-AR-transformer,diverse-inp-AR-transformer} rely on Auto-Regressive Models (AR models) \cite{RNN,attention-transformer}. Despite their remarkable success, these methods suffer from training instability and are typically limited to specialized degraded datasets, consequently exhibiting poor generalization to unseen masks and images.

\begin{figure}
\includegraphics[width=\textwidth]{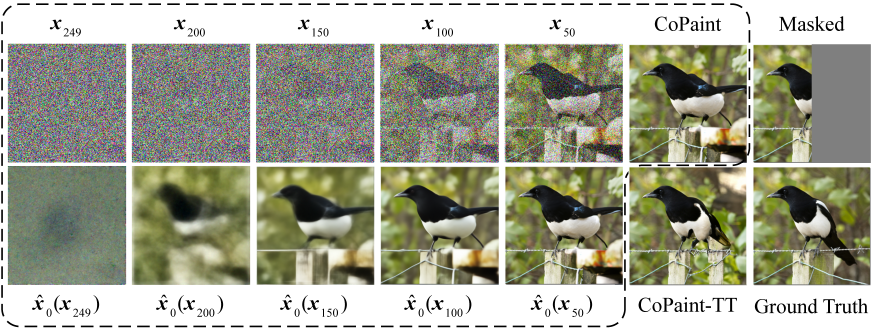}
\caption{When $T$=250, CoPaint's intermediate results show that flawed initial layouts lead to visually flawed outputs. For comparison, CoPaint-TT's refined generation is presented.} \label{fig2}
\end{figure}

\noindent The introduction of Denoising Diffusion Probabilistic Models (DDPM) \cite{DDPM} has elevated Diffusion Models (DMs) \cite{DPM} to state-of-the-art status in generative paradigms, demonstrating remarkable capability in producing high-fidelity and diverse images. Subsequent improvements \cite{improved-DDPM,DDIM,guided-diffusion,classifier-free,stablediffusion} to DMs have further enhanced their sampling efficiency and generation quality while extending their application to conditional generation, enabling tasks such as image-to-image generation \cite{palette}, text-to-image generation\cite{dalle2,stablediffusion,imagen}, video generation \cite{video-dm,video-dpm}, and 3D generation \cite{renderdiffusion}. Due to the powerful generative capabilities of diffusion models, an increasing number of studies have explored their application to image inpainting tasks, yielding remarkable results. Diffusion-based inpainting methods can be broadly categorized into two types: multimodal inpainting \cite{blended,blendedlatent,glide,SDEidt,smartbrush,diffedit,pbe,udifftext,structure} and context-aware inpainting \cite{repaint,mcg,ddrm,ddnm,dps,copaint,rad}. Multimodal methods typically require architectural modifications and model retraining, resulting in high computational costs, which makes them impractical for many real-world applications. In contrast, context-aware methods eliminate the need for retraining and offer much lower computational overhead, making them highly appealing. 

However, existing context-aware methods lack effective control over the missing regions, making it difficult to generate user-desired content, and when presented with incomplete targets, these methods struggles to generate visually plausible content, as illustrated in Fig.~\ref{fig1}. Owing to the stochastic nature of RePaint's \cite{repaint} intermediate generations, we only present CoPaint's \cite{copaint} intermediate results and their corresponding final predictions (see Fig.~\ref{fig2}). We observe that due to DDIM's \cite{DDIM} deterministic sampling, flawed global layouts generated in early denoising stages propagate through the process, resulting in visually implausible final outputs. The resampling extension (CoPaint-TT) not only demonstrates limited error correction capability but also results in significantly extended inference time. To address these limitations, we propose GuidPaint, and perform extensive evaluations on ImageNet \cite{imagenet} and CelebA-HQ \cite{celebahq} datasets. The contributions of our method are summarized as follows:
\begin{itemize}
    \item We propose a training-free class-guided inpainting algorithm that leverages a pretrained classifier to steer intermediate generations through both global and local guidance.  By optimizing intermediate outputs via gradient descent, we ensure the final predictions align with target classes. 
    \item We propose a hybrid sampling strategy combining stochastic and deterministic approaches.  This ensures semantically coherent and visually plausible outputs while preserving diversity. Crucially, our framework makes intermediate denoising steps visible to users, enabling interactive selection of preferred initial conditions for deterministic refinement.
\end{itemize}

\section{Related Work}
\subsubsection{Diffusion Models} DDPM \cite{DDPM} established the noise-perturbed Markov chain framework, formulating diffusion as a fixed-length variational autoencoder. iDDPM \cite{improved-DDPM} improved sample quality via learned noise schedules and hybrid objective functions. DDIM \cite{DDIM} introduced non-Markovian jumps enabling deterministic sampling with 10× speedup while preserving sample quality. Classifier Guidance (CG) \cite{guided-diffusion} leveraged gradient signals from pretrained classifiers to enhance controllability, albeit with trade-offs in diversity. Classifier-Free Guidance (CFG) \cite{classifier-free} eliminated the need for auxiliary classifiers through implicit conditional modeling, achieving better fidelity-diversity balance. Latent Diffusion Models (LDM) \cite{stablediffusion} operated in compressed latent spaces, reducing computational costs while maintaining high-resolution generation capabilities.

\begin{figure}
\includegraphics[width=\textwidth]{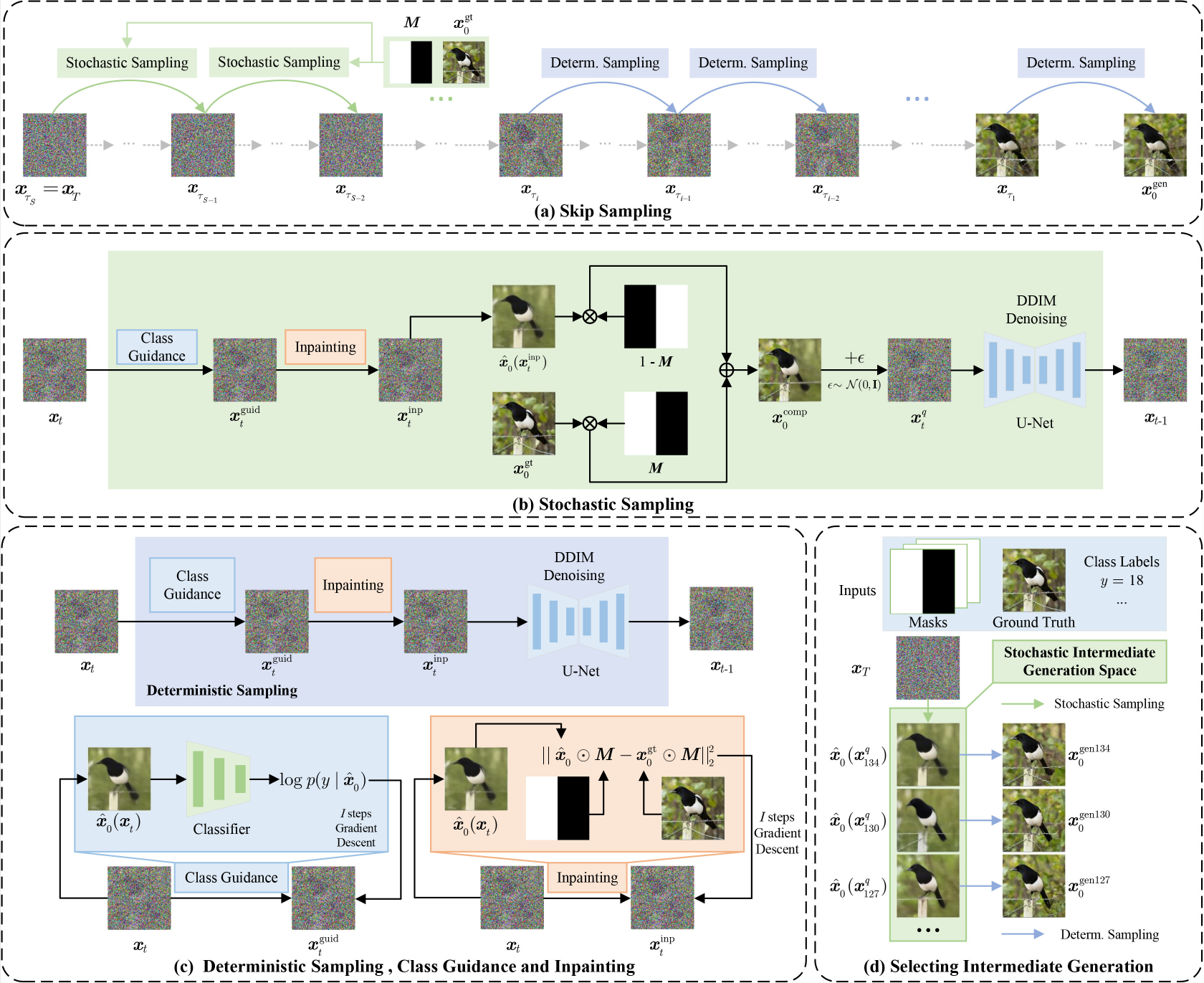}
\caption{GuidPaint overview. Input a ground truth image, masks and class labels. Skip sampling from random Gaussian noise, stochastic sampling generates diverse intermediate candidates, a selected candidate is then refined through deterministic sampling to produce the final output.} \label{fig3}
\end{figure}

\subsubsection{Context-Aware Image Inpainting} RePaint \cite{repaint} introduces DDPM \cite{DDPM} into image inpainting by concatenating noisy versions of known regions with intermediate generations of unknown regions, followed by multiple rounds of resampling, thus achieving semantically coherent and visually plausible results. CoPaint \cite{copaint}, which is based on DDIM \cite{DDIM}, avoids directly modifying the known regions. Instead, it performs iterative gradient-based optimization on intermediate generations to minimize the reconstruction error in the known regions, thereby enhancing the coherence between known and missing areas and ensuring greater visual consistency across the entire image. RAD \cite{rad} implements pixel-wise noise scheduling, enabling asynchronous generation of local regions while maintaining global contextual coherence.

\section{GuidPaint Method}
To ensure semantic consistency and visual plausiblility between the masked and unmasked regions, while enabling controllable generation within the masked areas, we propose GuidPaint. An overview of the proposed method is shown in Fig.~\ref{fig3}.
\subsection{Background}
DDPM \cite{DDPM} constructs a Markov chain-based forward diffusion process where, given an input image $\bm{x}_0$, Gaussian noise is progressively added through $T$ steps according to a linear noise schedule $\{\beta_t\}$. As $T \to \infty$, the resulting distribution converges to an isotropic Gaussian:

\begin{equation}
\begin{split}
    q(\bm{x}_t|\bm{x}_{t-1}) &= \mathcal{N}\left(\sqrt{1 - \beta_t}\,\bm{x}_{t-1}, \beta_t\mathbf{I}\right) \label{eq:transition}\\
    q(\bm{x}_{1:T}|\bm{x}_0) &= \prod_{t=1}^T q(\bm{x}_t|\bm{x}_{t-1}) 
\end{split}
\end{equation}

\noindent Through the reparameterization of $\beta_t$, the noisy sample $\bm{x}_t$ can be obtained in a single step from $\bm{x}_0$:

\begin{equation}
\begin{split}
    \bm{x}_t &= \sqrt{\alpha_t}\,\bm{x}_{t-1} + \sqrt{1-\alpha_t}\,\bm{\epsilon} \\
               &= \sqrt{\bar{\alpha}_t}\,\bm{x}_0 + \sqrt{1-\bar{\alpha}_t}\,\bm{\epsilon}
\end{split}
\label{eq:xt_addnoise}
\end{equation}

\noindent where $\alpha_t = 1 - \beta_t$, $\bar{\alpha}_t = \prod_{i=1}^{t} \alpha_i \,, \bm{\epsilon} \sim \mathcal{N}(\bm{0}, \mathbf{I})$. Therefore $q(\bm{x}_t|\bm{x}_0) = \mathcal{N}(\sqrt{\bar{\alpha}_t}\bm{x}_0,(1-\bar{\alpha}_t)\mathbf{I})$. DDPM employs a simplified variational bound objective for training:

\begin{equation}
    \mathcal{L} = \displaystyle \mathbb{E}_{\bm{x}_0, t, \bm{\epsilon}} \Vert \bm{\epsilon} - \bm{\epsilon}_\theta(\bm{x}_t, t) \Vert_2^2,
\end{equation}

\noindent where $\bm{\epsilon}_\theta$ is a noise estimator network based on U-Net \cite{unet}, $t \sim \mathcal{U}\{1, 2, \cdots, T\}$.

For the reverse denoising process, DDIM \cite{DDIM} introduces the inference distribution $q_\sigma$, where $\bm{x}_{t-1}$ is jointly determined by $\bm{x}_0$ and $\bm{x}_t$. Based on Bayes' theorem, DDIM derives a non-Markovian diffusion process, which enables skip sampling and significantly improves sampling efficiency.

\begin{equation}
\begin{gathered}
q_\sigma(\bm{x}_{1:T}|\bm{x}_0) = q_\sigma(\bm{x}_T|\bm{x}_0)\prod_{t=2}^T q_\sigma(\bm{x}_{t-1}|\bm{x}_t, \bm{x}_0) \\
q_\sigma(\bm{x}_{t-1}|\bm{x}_t,\bm{x}_0) = \mathcal{N}\left(\sqrt{\bar{\alpha}_{t-1}} \bm{x}_0 + \sqrt{1-\bar{\alpha}_{t-1}-\sigma_t^2} \frac{\bm{x}_t - \sqrt{\bar{\alpha}_t}\bm{x}_0}{\sqrt{1-\bar{\alpha}_t}}, \sigma_t^2 \mathbf{I}\right)
\end{gathered}
\label{eq:ddim_sampling}
\end{equation}

\noindent where $q_\sigma(\bm{x}_t|\bm{x}_0) = q(\bm{x}_t|\bm{x}_0), \forall t$. The denoising process initiates from a random Gaussian noise $\bm{x}_T \sim \mathcal{N}(\bm{0}, \mathbf{I})$. The transition equation of denoising process $p_\theta$ can be derived from $q_\sigma$ by simply replacing $\bm{x}_0$ with its estimate $\hat{\bm{x}}_0$: $p_\theta(\bm{x}_{t-1}|\bm{x}_t,\hat{\bm{x}}_0)$.
when $\sigma_t = 0$, the denoising process becomes deterministic. 

\begin{equation}
    \hat{\bm{x}}_0(\bm{x}_t) = \frac{\bm{x}_t-\sqrt{1-\bar{\alpha}_t}\bm{\epsilon}_\theta(\bm{x}_t,t)}{\sqrt{\bar{\alpha}_t}} \label{eq:predx0}
\end{equation}

\noindent CoPaint \cite{copaint} proposes an indirect optimization approach for modifying known regions in intermediate generations through iterative gradient descent, thereby ensuring coherence between known and unknown regions. We adapt their method as an inpainting constraint (Fig.~\ref{fig3}c), with the optimization objective defined as:

\begin{equation}
    \mathcal{L}_\text{inp}=\vert\vert \bm{x}_0^{\text{gt}} \odot \bm{M} - \hat{\bm{x}}_0 \odot \bm{M}\vert\vert_2^2 + \lambda_{\text{reg}}\vert\vert \bm{x}_t - \bm{\mu}_t\vert\vert_2^2 
\end{equation}

\noindent where $\bm{x}_0^{\text{gt}}$ denotes the ground truth image, $\bm{\mu}_t$ represents the intermediate generation before inpainting optimization at timestep $t$, and $\lambda_{\text{reg}}$ is a hyperparameter set to 0.01 by default.

\begin{equation}
    \bm{x}_t^{\text{inp}} = \bm{x}_t - \eta_t \nabla_{\bm{x}_t} \mathcal{L}_{\text{inp}}
\end{equation}

\noindent where the learning rate $\eta_t$ is set to  $0.02\sqrt{\bar{\alpha}_t}\cdot 1.012^{T-t}$ by default, with $1.012^{T-t}$ being a decay factor. $\bm{x}_t^{\text{inp}}$ is obtained after $I_{\text{inp}}$ steps of gradient descent, where $I_{\text{inp}}$ is set to 2.  

\subsection{Class-Conditional Guidance}
Classifier-Free Guidance(CFG) \cite{classifier-free} introduces a training strategy that probabilistically drops conditional inputs to co-train unconditional and conditional models within a single network, enabling controllable generation by linearly interpolating conditional and unconditional noise predictions. While effective for generative tasks, this approach struggles in inpainting tasks where prior knowledge of input images (e.g., class labels) is unavailable. Therefore, we adopt Classifier Guidance(CG) \cite{guided-diffusion}, leveraging a pretrained classifier to explicitly extract semantic information from inputs for restoration steering.

\subsubsection{Global Guidance}
The global guidance operates on the entire image when incomplete targets exist. We employ a pretrained classifier to predict Top-$k$ class labels from either ground truth ($\bm{x}_0^{\text{gt}}$) or masked ground truth ($\bm{x}_0^{\text{gt}} \odot \bm{M}$), even when the true labels are unknown, 
\begin{figure}
\includegraphics[width=\textwidth]{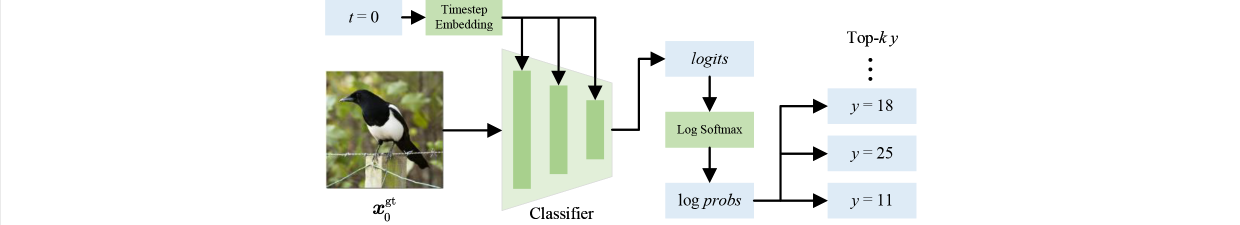}
\caption{Class label prediction using a pretrained classifier \cite{guided-diffusion}.} \label{fig4}
\end{figure}
as shown in Fig.~\ref{fig4}. Global guidance significantly enhances texture details in generated results while mitigating semantic inconsistencies (e.g., distorted object structures) and visual implausibilities (e.g., artifacts or blurring) during denoising. Similar to the inpainting constraints, we perform $I_{\text{guid}}$ steps gradient descent on $\bm{x}_t$ to minimize the cross-entropy loss, ensuring intermediate generations $\bm{x}_t$ adhere to the target class $y$ (Fig.~\ref{fig3}c):

\begin{equation}
\begin{gathered}
    \mathcal{L}_{\text{guid}} = - \log p_\phi(y|\hat{\bm{x}}_0) \label{eq:loss_guid}\\
    \bm{x}_t^{\text{guid}} = \bm{x}_t - s \bm{\mathrm{\Sigma}} \nabla_{\bm{x}_t} \mathcal{L}_{\text{guid}}
\end{gathered}
\end{equation}

\noindent where $s$ is the guidance scale (default=1.0, consistent with CG \cite{guided-diffusion}), and $\bm{\mathrm{\Sigma}}$ denotes the learnable reverse process variance proposed in Improved DDPM \cite{improved-DDPM}, parameterized as: 

\begin{equation}
    \bm{\mathrm{\Sigma}} = \exp\left( \bm{\upsilon} \log \beta_t + \left( 1 - \bm{\upsilon} \right) \log \tilde{\beta}_t \right)
\end{equation}

\noindent where $\bm{\upsilon} = (\bm{\mathrm{\Sigma}}_\theta(\bm{x}_t,t)+1)/2$, $\bm{\mathrm{\Sigma}}_\theta(\bm{x}_t,t)$ denotes the predicted variance produced by the neural network, $\tilde{\beta}_t$ is the variance in DDPM \cite{DDPM} corresponding to lower bounds for the reverse process variances.

\subsubsection{Local Guidance}
The local guidance focuses solely on masked regions for images without incomplete targets. Local guidance cannot be performed in a batch-processing manner and is limited to single-image inpainting, as it requires custom inputs of one or more masks along with the corresponding class labels for each mask. For all provided masks $\{\bm{M}_i\}_{i=1}^n$, local guidance is applied to each masked region sequentially by replacing $\hat{\bm{x}}_0$ in Eq.~\ref{eq:loss_guid} with $\hat{\bm{x}}_0 \odot (1-\bm{M}_i)$.

\subsection{Stochastic Sampling}
RePaint \cite{repaint} directly concatenates the noised version of known region with the intermediate generated content in missing region. However, the inherent noise mismatch between these components leads to semantic inconsistency and visual implausibilities. RePaint addresses this through iterative resampling, employing repeated stochastic noise corruption and reconstruction to alleviate such inconsistency. In contrast, our method composites the unmasked region $\bm{x}_0^{\text{gt}} \odot \bm{M}$ with the  masked region of $\hat{\bm{x}}_0$. The composited image $\bm{x}_0^{\text{comp}}$ is then forward-noised to the current timestep $t$ before DDIM denoising (Fig.~\ref{fig3}b).

\begin{equation}
\begin{gathered}
        \bm{x}_0^\text{comp} = \bm{x}_0^{\text{gt}} \odot \bm{M} + \hat{\bm{x}}_0 \odot (1-\bm{M}) \\
        \bm{x}_t^q = \sqrt{\bar{\alpha}_t}\,\bm{x}_0^\text{comp} + \sqrt{1-\bar{\alpha}_t}\,\bm{\epsilon}
\end{gathered}
\label{eq:comp_noise}
\end{equation}

\noindent where $\bm{\epsilon} \sim \mathcal{N}(\mathbf{0},\mathbf{I})$. As formalized in Eq.~\ref{eq:xt_addnoise}, the DDPM training process follows an identical paradigm of applying random noise to the complete image. This shared noise formulation ensures semantic coherence and prevents visual implausibilities between masked and unmasked regions during early and intermediate denoising phases. However, in later denoising steps, discrepancies between composited regions amplify, causing progressive divergence. We address this by introducing an early-stopping threshold $t_{\text{stop}}^{\text{comp}}$. We define the space composed of stochastically sampled intermediate results as the Stochastic Intermediate Generation Space, which maintains user-accessible visibility (Fig.~\ref{fig3}d). Users may select any desired intermediate generation as the initial condition for deterministic sampling (Fig.~\ref{fig3}c), following the formulation in Eq.~\ref{eq:ddim_sampling} with $\sigma_t = 0$.

\subsection{Non-Uniform Skip Sampling}
\begin{figure}
\includegraphics[width=\textwidth]{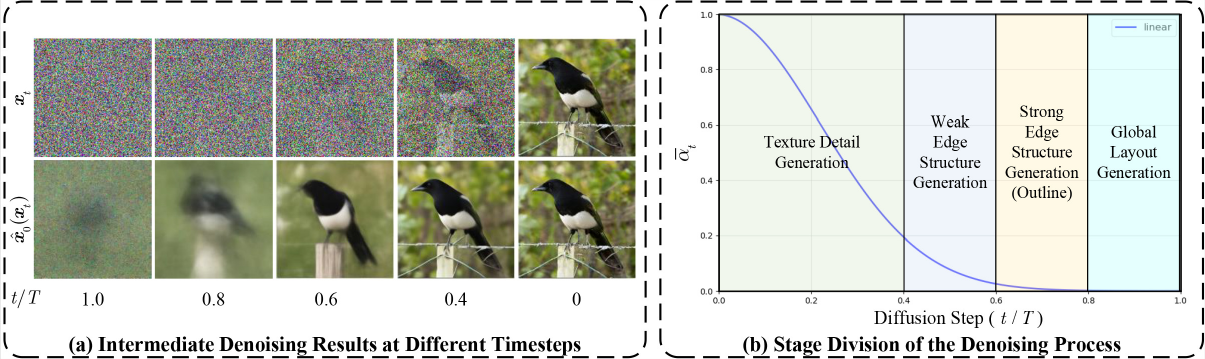}
\caption{(a) Visualization of intermediate denoising results at different timesteps; (b) Stage partitioning of the denoising process using a linear noise schedule.} \label{fig5}
\end{figure}
\noindent While DDIM \cite{DDIM} enables skip sampling, uniform skip intervals are suboptimal across the entire denoising process. As shown in Fig.~\ref{fig5}a, we partition the complete denoising process into four distinct stages, as visually summarized in Fig.~\ref{fig5}b. We define a stage-step set $\{s_i\}_{i=1}^n$, where $n$ indexes the $n$ partitions of the complete timestep sequence $\{1,2,\dots,T\}$ and $s_i$ determines the sampling steps for the $i$-th stage. The final skip-step sequence $\{\tau_i\}_{i=1}^S$ is constructed as a strictly increasing subsequence of $\{1,2,\dots,T\}$, where $S=\sum_{i=1}^{n}s_i$, $\tau_S=T$ (Fig.~\ref{fig3}a).

\section{Experiments}
\subsection{Experiment Setup}
\subsubsection{Datasets and DMs} We evaluate our method on both ImageNet \cite{imagenet} and CelebA-HQ \cite{celebahq} datasets. For evaluation, we choose 100 images from the validation set, resized to $256\times256$ resolution. For CelebA-HQ, we specifically select facial images with decorative accessories (e.g., sunglasses, hat). Hyperparameter selection is performed using the first 5 validation images. We use an unconditional DM pretrained on CelebA-HQ \cite{repaint}, and a conditional DM pretrained on ImageNet with a classifier \cite{guided-diffusion}. Following RePaint \cite{repaint} and CoPaint \cite{copaint}, we adopt three distinct mask types: Expand ($75\%$ masked), Half ($50\%$ masked), and Square ($25\%$ masked). Expand preserves only the central $128\times128$ region, Square masks the central $128\times128$ area while retaining the remaining regions, and Half keeps exclusively the left half of the image.
\subsubsection{Metics} For quantitative evaluation and comparison with existing methods, we employ three standard metrics: LPIPS (Learned Perceptual Image Patch Similarity) \cite{lpips}, PSNR (Peak Signal-to-Noise Ratio), and SSIM (Structural Similarity Index).
\subsubsection{Baselines and Implementation Details} For fair comparison with diffusion-based context-aware inpainting approaches, we select RePaint \cite{repaint} and CoPaint-TT \cite{copaint} as baselines, ensuring identical experimental conditions through using identical pretrained models and official code implementations with default settings. All methods use $T=250$ timesteps unless otherwise specified. For denoising, RePaint employs DDPM sampling, while both CoPaint-TT and GuidPaint (Ours) adopt DDIM sampling. For GuidPaint, we use $I_{guid}=I_{inp}=2$ and $t_{stop}^{comp}=130$. For GuidPaint-Skip, we set step counts $\{50,50,25,25,5\}$ for skip sampling, $I_{guid}=I_{inp}=1$ and $t_{stop}^{comp}=124$. we use guidance scale $s=1.0$ for all classifier guidance (CG) and use global guidance. All experiments were conducted on NVIDIA RTX 3090 GPU.
\subsection{Experiment Results}
Tab.~\ref{tab1} presents quantitative results, demonstrating our method's superior performance on large-area masks (Expand and Half). On Square mask, it achieves significantly lower error and higher structural similarity. For CelebA-HQ, our stochastic sampling (SS) further reduces perceptual discrepancy while enhancing structural consistency. Tab.~\ref{tab2} shows sampling efficiency. Our non-uniform skip sampling accelerates inference  without quality degradation.

\begin{table}[]
\centering
\caption{ImageNet (top) and CelebA-HQ (bottom) Quantitative Results. $^\ast$ indicates that GuidPaint uses unconditional DM without CG.}
\label{tab1}
\begin{tabular}{lccccccccc}
\toprule
\textbf{ImageNet}  & \multicolumn{3}{c}{Expand}                                   & \multicolumn{3}{c}{Half}                                     & \multicolumn{3}{c}{Square}                                   \\ \cmidrule(lr){2-4} \cmidrule(lr){5-7} \cmidrule(lr){8-10} 
Method             & \multicolumn{1}{c}{LPIPS${\downarrow}$} & \multicolumn{1}{c}{PSNR${\uparrow}$} & SSIM${\uparrow}$ & \multicolumn{1}{c}{LPIPS${\downarrow}$} & \multicolumn{1}{c}{PSNR${\uparrow}$} & SSIM${\uparrow}$ & \multicolumn{1}{c}{LPIPS${\downarrow}$} & \multicolumn{1}{c}{PSNR${\uparrow}$} & SSIM${\uparrow}$ \\ \midrule
RePaint\cite{repaint}            & \multicolumn{1}{c}{0.462}      & \multicolumn{1}{c}{12.271}     &0.488      & \multicolumn{1}{c}{0.296}      & \multicolumn{1}{c}{14.821}     &0.655      & \multicolumn{1}{c}{0.165}      & \multicolumn{1}{c}{18.492}     &\textbf{0.814}      \\ 
CoPaint-TT\cite{copaint}         & \multicolumn{1}{c}{0.427}      & \multicolumn{1}{c}{12.016}     &0.460      & \multicolumn{1}{c}{0.278}      & \multicolumn{1}{c}{14.171}     &0.631      & \multicolumn{1}{c}{0.171}      & \multicolumn{1}{c}{18.335}     &0.804      \\ \midrule
GuidPaint-Skip     & \multicolumn{1}{c}{0.409}      & \multicolumn{1}{c}{13.988}     &0.507      & \multicolumn{1}{c}{0.265}      & \multicolumn{1}{c}{15.826}     &0.657      & \multicolumn{1}{c}{0.156}      & \multicolumn{1}{c}{\textbf{19.601}}     &0.802 \\ 
GuidPaint          & \multicolumn{1}{c}{\textbf{0.340}}      & \multicolumn{1}{c}{\textbf{14.532}}     &\textbf{0.532}      & \multicolumn{1}{c}{\textbf{0.262}}      & \multicolumn{1}{c}{\textbf{15.894}}     &\textbf{0.657}      & \multicolumn{1}{c}{\textbf{0.155}}      & \multicolumn{1}{c}{19.388}     &0.805      \\ \toprule
\textbf{CelebA-HQ} & \multicolumn{3}{c}{Expand}                                   & \multicolumn{3}{c}{Half}                                     & \multicolumn{3}{c}{Square}                                   \\ \cmidrule(lr){2-4} \cmidrule(lr){5-7} \cmidrule(lr){8-10} 
Method             & \multicolumn{1}{c}{LPIPS${\downarrow}$} & \multicolumn{1}{c}{PSNR${\uparrow}$} & SSIM${\uparrow}$ & \multicolumn{1}{c}{LPIPS${\downarrow}$} & \multicolumn{1}{c}{PSNR${\uparrow}$} & SSIM${\uparrow}$ & \multicolumn{1}{c}{LPIPS${\downarrow}$} & \multicolumn{1}{c}{PSNR${\uparrow}$} & SSIM${\uparrow}$ \\ \midrule
RePaint\cite{repaint}            & \multicolumn{1}{c}{0.401}      & \multicolumn{1}{c}{10.682}     &0.507      & \multicolumn{1}{c}{0.215}      & \multicolumn{1}{c}{14.521}     &0.705      & \multicolumn{1}{c}{0.096}      & \multicolumn{1}{c}{20.933}     &\textbf{0.862}      \\ 
CoPaint-TT\cite{copaint}         & \multicolumn{1}{c}{\textbf{0.362}}      & \multicolumn{1}{c}{11.134}     &0.517      & \multicolumn{1}{c}{0.203}      & \multicolumn{1}{c}{14.848}     &0.694      & \multicolumn{1}{c}{\textbf{0.090}}      & \multicolumn{1}{c}{21.579}     &0.849      \\ \midrule
GuidPaint-Skip$^{\ast}$    & \multicolumn{1}{c}{0.386}      & \multicolumn{1}{c}{12.061}     &0.541      & \multicolumn{1}{c}{0.214}      & \multicolumn{1}{c}{16.123}     &0.711      & \multicolumn{1}{c}{0.112}      & \multicolumn{1}{c}{\textbf{21.811}}     &0.831      \\ 
GuidPaint$^{\ast}$         & \multicolumn{1}{c}{0.367}      & \multicolumn{1}{c}{\textbf{12.397}}     &\textbf{0.557}      & \multicolumn{1}{c}{\textbf{0.201}}      & \multicolumn{1}{c}{\textbf{16.216}}     &\textbf{0.725}      & \multicolumn{1}{c}{0.096}      & \multicolumn{1}{c}{21.798}     &0.845      \\ \bottomrule
\end{tabular}
\end{table}

\begin{table}[]
\begin{minipage}{0.48\textwidth}
\centering
\caption{Sampling time comparison (ImageNet, $1\times$ NVIDIA RTX 3090)}
\label{tab2}
\begin{tabular}{lc}
\toprule
Method         & time(s) \\ \midrule
RePaint \cite{repaint}        &183         \\ 
CoPaint-TT \cite{copaint}     &330         \\ \midrule
GuidPaint-Skip &\textbf{138}         \\ 
GuidPaint       &381         \\ 
\bottomrule        
\end{tabular}
\end{minipage}
\hfill  
\begin{minipage}{0.48\textwidth}
\centering
\caption{Ablation study on ImageNet with Half mask. Where CG is Classifier Guidance and SS is Stochastic Sampling.}
\label{tab3}
\begin{tabular}{lccc}
\toprule
Method    & LPIPS${\downarrow}$ & PSNR${\uparrow}$ & SSIM${\uparrow}$ \\ \midrule
w/o CG    &0.265       &15.551      &0.643      \\
w/o SS    &0.278       &14.903      &0.634      \\
GuidPaint &\textbf{0.262}       &\textbf{15.894}      &\textbf{0.657 }     \\
\bottomrule   
\end{tabular}
\end{minipage}
\end{table}

\begin{figure}
\includegraphics[width=\textwidth]{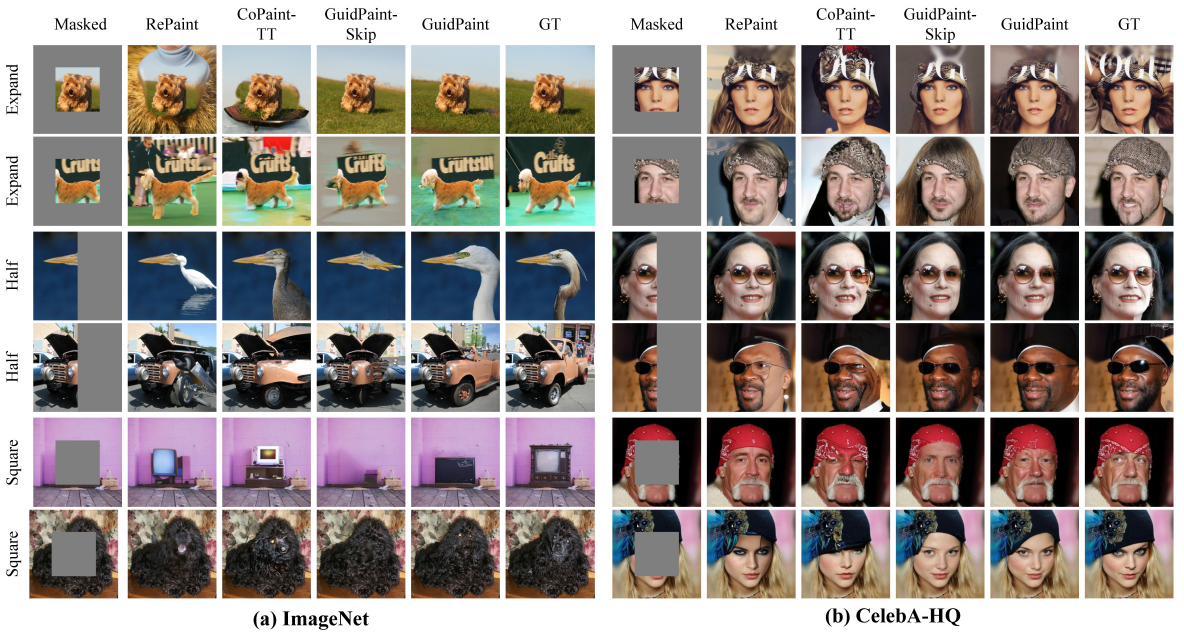}
\caption{Qualitative results of ImageNet (a) and CelebA-HQ (b).} \label{fig6}
\end{figure}

\begin{figure}
\includegraphics[width=\textwidth]{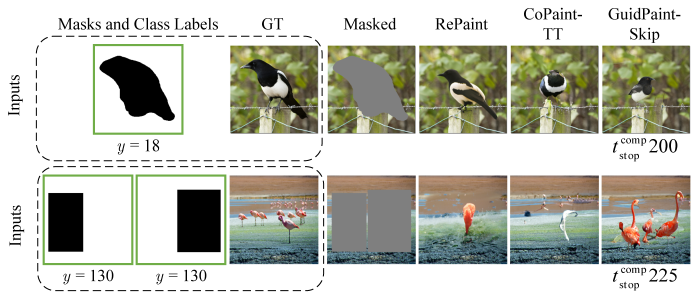}
\caption{Qualitative results of Local Guidance on ImageNet (18: ``magpie", 130: ``flamingo").} \label{fig7}
\end{figure}

Fig.~\ref{fig6} shows qualitative comparisons. Our method exhibits better semantic coherence and visual plausibility, contrasting with RePaint's and CoPaint-TT's unresonable results. Fig.~\ref{fig7} shows the results of Local Guidance. It can be seen that after applying classifier guidance, the desired results can be generated in masked regions.

\subsection{Ablation Study}
\begin{figure}
\centering
\includegraphics[width=0.7\textwidth]{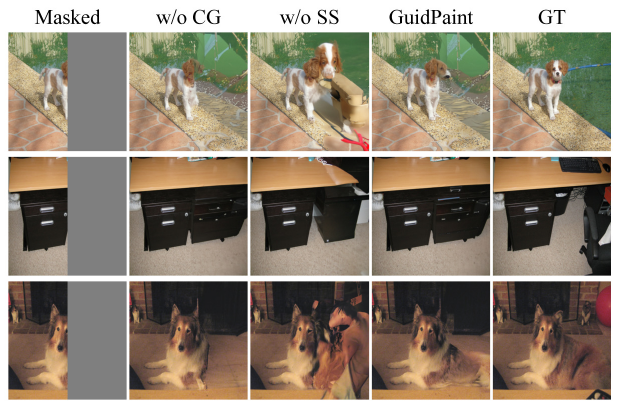}
\caption{Ablation study examples on ImageNet with Half mask (CG: Classifier Guidance, SS: Stochastic Sampling).} \label{fig8}
\end{figure}

\begin{figure}
\includegraphics[width=\textwidth]{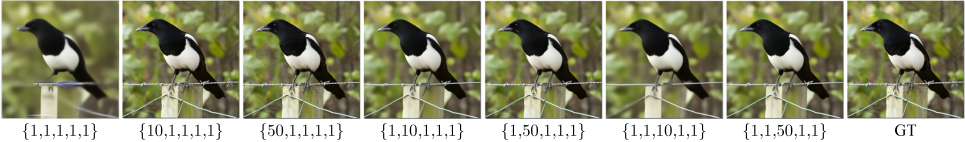}
\caption{Ablation study examples on ImageNet with Skip Sampling.} \label{fig9}
\end{figure}

Fig.~\ref{fig8} shows the ablation study results, demonstrating that stochastic sampling effectively mitigates implausible intermediate generations, while classifier guidance reduces semantic deviations. Fig.~\ref{fig9} demonstrates that the texture refinement stage requires the largest timestep allocation, while other stages can achieve optimal results with fewer steps.

\section{Limitations}
Our method currently supports only single-class multi-instance generation due to the inability to parallelize conditional model calls across different class labels. The framework accepts only class labels as conditional inputs, thus excluding text prompts or segmentation maps.
\section{Conclusions}
We propose a training-free class-guided inpainting framework that leverages a pretrained classifier for global and local guidance while introducing skip sampling to accelerate generation. Through gradient-based optimization of intermediate outputs, our method ensures both class consistency and high-fidelity results. The hybrid sampling strategy synergizes stochastic exploration with deterministic refinement, maintaining semantic coherence and visual plausibility without compromising diversity. Notably, our interactive pipeline reveals intermediate denoising steps with skip sampling visualization, enabling users to select optimal initial conditions for deterministic generation.



\bibliographystyle{splncs04}
\bibliography{refs}

\end{document}